\documentclass{article}
\usepackage[preprint]{corl_2023} 
\usepackage{times}
\usepackage{epsfig}
\usepackage{graphicx}
\usepackage{amsmath}
\usepackage{amssymb}
\usepackage{booktabs}
\usepackage{multirow}
\usepackage{tabularx}
\usepackage{enumitem}
\usepackage{gensymb}
\usepackage{colortbl}
\usepackage{caption}
\usepackage{makecell}
\usepackage[table]{xcolor}
\usepackage[numbers]{natbib}
\usepackage{wrapfig}
\usepackage{amssymb}
\usepackage{pifont}
\definecolor{mycyan}{cmyk}{.1,0,0,0}
\newcommand{\cmark}{\ding{51}}%
\newcommand{\cmarkg}{\textcolor{lightgray}{\ding{51}}}%
\newcommand{\xmark}{\ding{55}}%
\newcommand{\xmarkg}{\textcolor{lightgray}{\ding{55}}}%

\newcommand{\name}{LiDAR-NeRF}
\newcommand{\datasetname}{NeRF-MVL}
\newcommand{\ang}[1]{#1\degree}

\newcommand{\mypara}[1]{\vspace{1mm}\noindent\textbf{#1}}
\definecolor{amber}{rgb}{1.0, 0.49, 0.0}

\usepackage[capitalize]{cleveref}
\crefname{section}{Sec.}{Secs.}
\Crefname{section}{Section}{Sections}
\Crefname{table}{Table}{Tables}
\crefname{table}{Tab.}{Tabs.}

\begin{document}

\title{\name{}: Novel LiDAR View Synthesis via Neural Radiance Fields}

\author{
  Tang Tao$^{1}$\footnotemark[1] \quad
  Longfei Gao$^{2}$ \quad 
  Guangrun Wang$^{3}$ \quad
  Yixing Lao $^{4}$ \quad
  Peng Chen$^{2}$ \quad \\
  \textbf{
  Hengshuang Zhao $^{4}$ \quad
  Dayang Hao$^{2}$ \quad
  Xiaodan Liang$^{1}$\footnotemark[4] \quad
  Mathieu Salzmann$ ^{5}$ \quad
  Kaicheng Yu$^{2}$ \quad}
  \\
\fontsize{10pt}{\baselineskip}\selectfont
$^1$ Shenzhen Campus, Sun Yat-sen University $^2$ Autonomous Driving Lab, Alibaba Group\\
\fontsize{10pt}{\baselineskip}\selectfont
$^3$  University of Oxford  $^4$ The University of Hong Kong  $^5$ CVLab, EPFL \\
 \begin{normalsize}${\tt \{trent.tangtao, kaicheng.yu.yt,hxdliang328\}@gmail.com} $\end{normalsize}
}

{%
\maketitle

\vspace{-10pt}
\begin{center}
    \centering
    \captionsetup{type=figure}
    \begin{center}
        \includegraphics[width=1\linewidth]{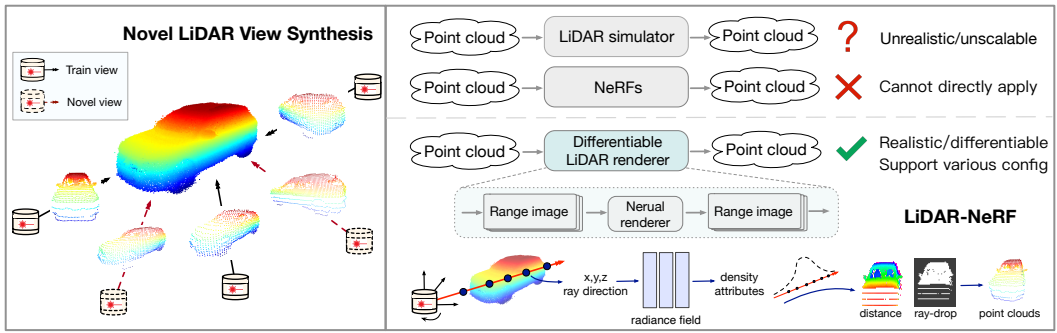}
    \end{center}
    \captionof{figure}{
        \textbf{(Left)} We introduce the task of novel view synthesis for LiDAR sensors. Given multiple LiDAR viewpoints of an object, novel LiDAR view synthesis aims to render a point cloud of the object from an arbitrary new viewpoint.
        \textbf{(Right, Top)} The mostly-closed related approaches to generating new LiDAR point clouds are some LiDAR simulators, which suffer from limited scalability and  applicability, and fails to produce realistic LiDAR patterns.
        Furthermore, traditional NeRFs are not directly applicable to point clouds.
        \textbf{(Right, Bottom)} By contrast, we propose a novel differentiable framework, \name{}, with an associated neural radiance field, to avoid explicit 3D reconstruction and game engine usage. Our method enables end-to-end optimization and encompasses the 3D point attributes
        into the learnable field.
    }
    \vspace{-0.5cm}
    \label{fig:teaser}
\end{center}%
}


\renewcommand{\thefootnote}{\fnsymbol{footnote}}
\footnotetext[1]{Work done during an internship at DAMO Academy, Alibaba Group.}
\footnotetext[2]{Corresponding Author.}
\renewcommand{\thefootnote}{\arabic{footnote}}

\begin{abstract}
    We introduce a new task, novel view synthesis for LiDAR sensors. While traditional model-based LiDAR simulators with style-transfer neural networks can be applied to render novel views, they fall short of producing accurate and realistic LiDAR patterns because the renderers rely on explicit 3D reconstruction and exploit game engines, that ignore important attributes of LiDAR points.
    We address this challenge by formulating, to the best of our knowledge, the first differentiable end-to-end LiDAR rendering framework, \name{}, leveraging a neural radiance field~(NeRF) to facilitate the joint learning of geometry and the attributes of 3D points. However, simply employing NeRF cannot achieve satisfactory results, as it only focuses on learning individual pixels while ignoring local information, especially at low texture areas, resulting in poor geometry. To this end, we have taken steps to address this issue by introducing a structural regularization method to preserve local structural details.
    To evaluate the effectiveness of our approach, we establish an object-centric
    \textbf{m}ulti-\textbf{v}iew \textbf{L}iDAR dataset, dubbed \datasetname{}. It contains observations of objects from 9 categories seen from 360-degree viewpoints captured with multiple LiDAR sensors. Our extensive experiments on the scene-level KITTI-360 dataset, and on our object-level \datasetname{} show that our \name{} surpasses the model-based algorithms significantly. \href{https://tangtaogo.github.io/lidar-nerf-website/}{Our project page}.

\end{abstract}

\section{Introduction}
\begin{figure*}[h]
    \centering
    \vspace{-0.3cm}
    \includegraphics[width=1\linewidth]{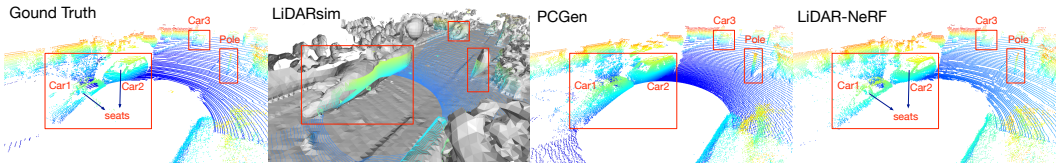}
    \vspace{-0.5cm}
    \caption{
        A comparison of novel view LiDAR point clouds generated from LiDARsim~\cite{manivasagam2020lidarsim}, PCGen~\cite{li2022pcgen}, and our \name{}. LiDARsim suffers from inaccuracies in explicit 3D mesh reconstruction. PCGen overestimates object surfaces. Specifically, laser beams emitted by the LiDAR sensor can be influenced by surface material and normal direction, resulting in some beams penetrating car glass and reaching the seats (\textbf{\color{purple}car1 and car2}), while others are lost (\textbf{\color{purple}car3}). Although an additional style-transfer net can alleviate the problem of beam loss, it does not take into account special attributes like the transmission. As opposed to prior arts, our proposed method, \name{}, effectively encodes 3D information and multiple attributes, achieving high fidelity with ground truth. We encourage readers to zoom in for better observations.
    }
    \vspace{-0.3cm}
    \label{fig:motivation}
\end{figure*}

\vspace{-4pt}
Synthesizing novel views of a scene from a given camera has been a longstanding and prominent subject of research. A recent milestone in this area has been to combine differentiable rendering with neural radiance fields (NeRF)~\cite{mildenhall2021nerf}, resulting in a de-facto standard to render photo-realistic novel views by leveraging only a hundred or fewer input images with known camera poses. Impressively, this has already been shown to positively impact downstream tasks such as autonomous driving~\cite{ost2021NSG,rematas2022urban-nerf,tancik2022blocknerf, xie2023s-nerf}.
In such an autonomous driving scenario, however, practical systems typically exploit not only images but also LiDAR sensors, which provide reliable 3D measurements of the environment. As such, it seems natural to seek to generate novel views not only in the image domain but also in the LiDAR one.
However, the only methods that consider LiDAR point clouds for novel view synthesis~\cite{rematas2022urban-nerf, xie2023s-nerf} only do so to boost training, thus still producing images as output.
In other words, generating novel LiDAR views remains unexplored.
Despite the 3D nature of this modality, this task remains challenging, as LiDARs only provide a partial view of the scene,
corrupted by various attributes related to the LiDAR physical modeling.
On the other hand, the mostly-closed related approaches
are some LiDAR simulators~\cite{manivasagam2020lidarsim, li2022pcgen}, which adopt a multi-step approach that reconstructs a 3D mesh from the input point clouds and utilizes game engines to simulate a new point cloud. Nevertheless, the intricacy of this approach can limit its practicality and scalability. Moreover, as shown in \cref{fig:motivation}, this strategy tends to produce unrealistic LiDAR patterns, as its explicit reconstruction and ray-casting overlook certain crucial features of LiDAR points.

In this paper, we hereby present the pioneering differentiable rendering method for novel LiDAR view synthesis. Unlike RGB view synthesis, the output of a free viewpoint LiDAR sensor is a point cloud sampled from the surrounding 3D scene according to the given LiDAR sensor-specific pattern, as illustrated in \cref{fig:teaser}. Consequently, the direct application of the NeRF formalism which relies on a photometric loss, is infeasible in this context.
To overcome this challenge, we first convert point clouds with respect to a surface plane, serving as a 360-degree range pseudo image in which each pseudo pixel represents the distance between the LiDAR receiver and a world point hit by a laser beam. We then use a neural network to encode the 3D information and predict multiple attributes for each pseudo-pixel. Specifically, we regress the distance of each pseudo pixel, which represents their 3D coordinate, its intensity, which encodes the amount of reflected light that reaches the sensor at a pseudo pixel, and an attribute that we dub ray-drop, which encodes the probability of dropping a pseudo pixel. This last attribute reflects our observation that, in the real world, some laser beams of the LiDAR sensor are simply lost, due to the surface material and normal direction.
As image-based NeRFs, our \name{} leverages multi-view consistency, thus enabling the network to produce accurate geometry. 
Despite these efforts, the performance remains suboptimal, as NeRFs concentrate solely on learning individual pixels while neglecting local information, particularly in low-texture regions of large-scale scenes, resulting in subpar geometry. To address this issue, we propose a structural regularization to preserve local structural details, which in turn serve as a guide for NeRFs geometry to produce more accurate estimations.

Validating the effectiveness of our approach can be achieved by leveraging the existing autonomous driving datasets~\cite{liao2022kitti360, caesar2020nuscenes, sun2020waymo} that provide LiDAR data. However, as these datasets were acquired from a vehicle moving along the street, the objects they depict are observed with limited viewpoint variations, thus making them best suited for scene-level synthesis. This contrasts with object-level synthesis, as is more common in image novel-view synthesis~\cite{mildenhall2021nerf, liu2020nsvf, knapitsch2017tanks}. We, therefore, establish an object-centric \textbf{m}ulti-\textbf{v}iew \textbf{L}iDAR dataset, which we dub the \datasetname{} dataset, containing carefully calibrated sensor poses, acquired from multi-LiDAR sensor data from real autonomous vehicles. It contains more than 76k frames covering two types of collecting vehicles, three LiDAR settings, two collecting paths, and nine object categories.

We evaluate our model’s scene-level and object-level synthesis ability on scenes from the challenging KITTI-360 dataset~\cite{liao2022kitti360} and from our \datasetname{} dataset both quantitatively and qualitatively.
Our results demonstrate the superior performance of our approach compared to the baseline renderer in various metrics and visual quality, showcasing its effectiveness in LiDAR novel view synthesis.


Overall, we make the following contributions: (1) We formulate the first differentiable framework, \name{}, for novel LiDAR view synthesis, which can render novel point clouds with point intensity and ray-drop probability without explicit 3D reconstruction.
(2) We propose a structural regularization method to effectively preserve local structural details, thereby guiding the model towards more precise geometry estimations, leading to more faithful novel LiDAR view synthesis.
(3) We establish the \datasetname{} dataset from LiDAR sensors of real autonomous vehicles to evaluate the object-centric novel LiDAR view synthesis.
(4) We demonstrate the effectiveness of our \name{}  quantitatively and qualitatively in both scene-level and object-level novel LiDAR view synthesis.

\vspace{-8pt}

\section{Related Work}
\vspace{-7pt}
\mypara{Novel RGB view synthesis.}
Synthesizing novel RGB views of a scene from a set of captured images is a long-lasting problem.
In particular, recent advances in NeRF have demonstrated their superior performance in synthesizing images, thanks to the pioneering work of NeRF~\cite{mildenhall2021nerf}.
Following this, many NeRF strategies have been proposed for acceleration~\cite{muller2022instantNGP, yu2021plenoxels, chen2022tensorf} and generalization~\cite{yu2021pixelnerf, kim2022infonerf, niemeyer2022regnerf}.
Noticeably, notions of depth have been used for novel RGB view synthesis~\cite{deng2022dsnerf, neff2021donerf, roessle2022depthpriors}.
In parallel, great progress has been made to handle complex environments, such as large-scale outdoor scenes~\cite{rematas2022urban-nerf, tancik2022blocknerf, xie2023s-nerf}, demonstrating the tremendous potential of NeRFs for real-world applications.
Nevertheless, while these works improve quality and convergence speed, they still produce RGB images.
In practical scenarios where multiple sensors, such as RGB cameras and LiDARs, are used, only synthesizing the image view is insufficient. In this work, drawing inspiration from NeRF~\cite{mildenhall2021nerf}, we introduce the first differentiable framework for novel LiDAR view synthesis.

\vspace{-4pt}
\mypara{Model-based LiDAR simulators.}
There are model-based LiDAR simulators that can also be regarded as LiDAR renderers.
In this context, early works~\cite{deschaud2021kitti-carla, yue2018lidar-GTA, wu2018squeezeseg-GTA, xiao2022transfer-UE4} employ graphics engines, such as CARLA~\cite{dosovitskiy2017carla}
, to simulate LiDAR sensors.
However, this yields a large sim-to-real domain gap, as their virtual worlds use handcrafted 3D assets and make simplified physics assumptions.
More recent works, e.g., LiDARsim~\cite{manivasagam2020lidarsim} and PCGen~\cite{li2022pcgen}, employ a multi-step, data-driven approach to simulate point clouds from real data.
They first leverage real data to reconstruct the 3D scene,
and then utilize the reconstructed 3D scene to render novel LiDAR data via ray-casting.
To close the sim-to-real gap, they further train a network to model the physics of LiDAR ray-dropping.
However, the multiple steps involved in this approach affect its applicability and scalability.
Additionally, this approach typically fails to generate authentic LiDAR patterns since its explicit reconstruction and ray-casting disregard some crucial attributes of LiDAR points.
By contrast, as the first differentiable LiDAR renderer, our approach is simple and effective, yet produces realistic LiDAR data.

\vspace{-6pt}

\section{Novel LiDAR View Synthesis}
In this section, we first give a formal problem definition of novel LiDAR view synthesis, and introduce our \name{} in detail. Finally, we describe our object-level multi-view LiDAR dataset.

\mypara{Problem definition.}
Novel LiDAR view synthesis aims to render an object or scene from an arbitrary new viewpoint given a set of existing observations acquired from other viewpoints. Formally, given a set $\mathcal{D} = \{(P_i, G_i)\}$, where $P_i$ is the LiDAR pose and $G_i$ is the corresponding observed point cloud, we aim to define a rendering function $f$ that can generate a new point cloud from an arbitrary new pose $P'$, i.e., $G' = f_{\mathcal{D}}(P')$.
\subsection{LiDAR Model and Range Representation}
\label{subsec:rangview}
\begin{figure}[t]
    \begin{center}
        \includegraphics[width=1\linewidth]{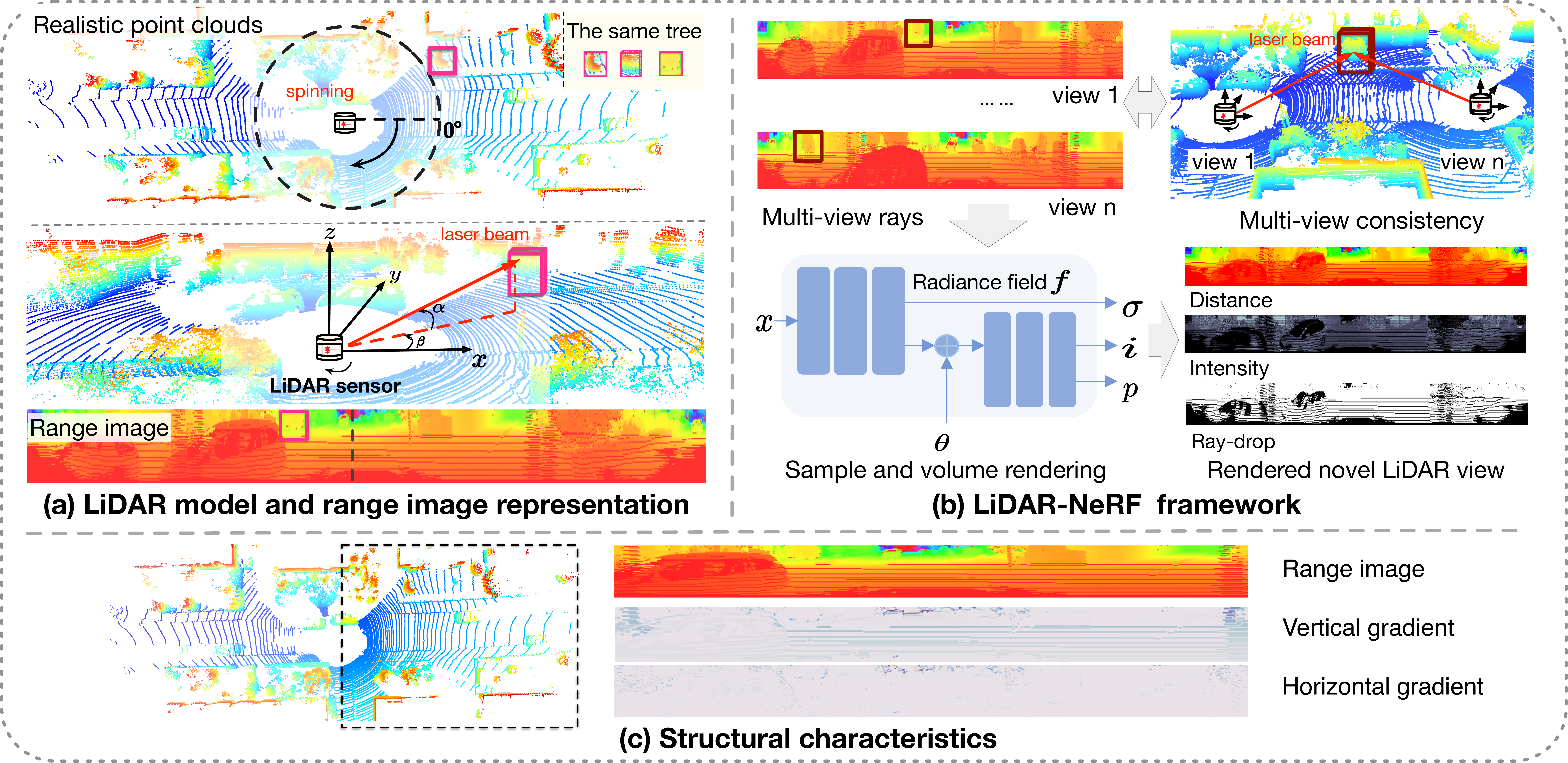}
    \end{center}
    \vspace{-4pt}
    \caption{ \textbf{(a)}
        \textbf{(Top)} The physical model of a LiDAR can be described as follow: each laser beam originates from the sensor origin and shoot outwards to a point in the real world or vanishes. One common pattern of laser beams is spinning in a 360-degree fashion. \textbf{(Bottom)} We convert the point clouds into a range image, where each pixel corresponds to a laser beam. Note that we highlight one object in the different views to facilitate the visualization.
        \textbf{(b)} Taking multi-view LiDAR range images with associated sensor poses as input, our model produces 3D representations of the distance, the intensity, and the ray-drop probability at each pseudo-pixel. We exploit multi-view consistency of the 3D scene to help our network produce accurate geometry.
        \textbf{(c)} The physical nature of LiDAR models results in point clouds exhibiting recognizable patterns, such as the ground primarily appearing as continuous straight lines. This pattern is also evident in the transformed range images, which display significant structural features, such as their horizontal gradient being almost zero in flat areas. As a result, these structural characteristics are essential for the network to learn.
    }
    \label{fig:LiDAR-NeRF}
    \vspace{-10pt}
\end{figure}

To produce accurate and realistic novel LiDAR views, we draw inspiration from the NeRF formalism.
However, one cannot directly apply the traditional NeRFs, which leverage a per-pixel photometric error measure, to novel LiDAR view synthesis, where the observations are 3D points.
To address this, we investigate the LiDAR model and convert point clouds into a range representation.
Let us start with the LiDAR model as shown in the top of \cref{fig:LiDAR-NeRF}~(a), which works by emitting a laser beam and measuring the time it takes for the reflected light to return to the sensor. For a LiDAR with $H$ laser beams in a vertical plane and $W$ horizontal emissions, the returned attributes (e.g., distance $d$ and intensity $i$) form an $H \times W$ range pseudo image.
Specifically, for the 2D coordinates $(h, w)$ in the range pseudo image, we have
\begin{align}
  \begin{pmatrix}
    \alpha \\
    \beta
  \end{pmatrix}
  =
  \begin{pmatrix}
    |f_{\text{up}}| - h f_{v} H^{-1} \\
    - \left( 2w - W \right)  \pi W ^{-1}
  \end{pmatrix},
  \label{eq:theta2}
\end{align}
where $f_{v} = |f_{\text{down}}| + |f_{\text{up}}|$ is the vertical field-of-view of the LiDAR sensor.
Conversely, each 3D point $(x, y, z)$ in a LiDAR frame can be projected on a range pseudo image of size $H \times W$ as
  {\footnotesize
    $
      \begin{pmatrix}
        h \\
        w
      \end{pmatrix}
      =
      \begin{pmatrix}
        \left( 1 - (\arcsin (z, d) + |f_{\text{down}}|) f_{v}^{-1} \right) H \\
        \frac{1}{2} \left( 1 - \arctan (y, x) \pi^{-1} \right) W
      \end{pmatrix}.
      \label{eq:rangeview}
    $
  }
Note that if more than one point projects to the same pseudo-pixel, only the point with the smallest distance is kept. The pixels with no projected points are filled with zeros.
In addition to the distance, the range image can also encode other point features, such as intensity.


\subsection{\name{} Framework}
\label{subsec:lidar-nerf}
Motivated by the impressive results of NeRF~\cite{mildenhall2021nerf} for novel RGB view synthesis, we therefore introduce the first  differentiable novel LiDAR view synthesis framework.

\vspace{-4pt}
\mypara{Implicit fields to represent LiDAR sensor.}
As discussed in \cref{subsec:rangview}, LiDAR sensors use an active imaging system that differs from the passive imaging principle of cameras, requiring specific modeling of the sensor's characteristics. Additionally, each pseudo pixel in the LiDAR range image corresponds to a real laser beam, which is more consistent with the rays in NeRF. Therefore, we reformulated NeRF to achieve novel LiDAR view synthesis. For a given LiDAR range image, the laser's viewing directions $\boldsymbol{\theta}$ of a pseudo pixel can be calculated using \cref{eq:theta2}. The viewing directions in our proposed LiDAR-NeRF framework form a radial pattern, closely matching physical reality as depicted in \cref{fig:LiDAR-NeRF} (a).
The expected depth can be obtained by integrating over samples as $
  \hat{D}(\textbf{r}) = \sum_{i=1}^N T_i \big(1 - \exp(-\sigma_i \delta_i)\big)t_i \;.
  \label{eq:depth}
$ where $ T_i = \exp \left(-\sum_{j=1}^{i-1} \sigma_j \delta_j \right)$ indicates the accumulated transmittance along ray $\textbf{r}$ to the sampled point $t_i$, and $\delta_i = t_{i+1} - t_i$ is the distance between adjacent samples.
The expected depth represents the distance from the LiDAR sensor, which is also represented by a pseudo pixel in the range image. Moreover, both the origin $\mathbf{o}$ and viewing direction $\boldsymbol{\theta}$ of the ray are transformed to the global world coordinate system, allowing the sampled points $t$ to be actual points in the real world and consistent across multiple LiDAR frames/range images, as shown in the top-right portion of \cref{fig:LiDAR-NeRF} (b).

With these geometry aspects in mind, we developed a framework that can: 1) synthesize a novel LiDAR frame with realistic geometry; 2) estimate LiDAR intensities over the scene; and 3) predict a binary ray-drop mask that specifies where rays will be dropped. Both ray-drop and intensity can be recorded as color features of LiDAR~\cite{park2017colored}, and both are view-dependent.
For the radiance field, we follow the traditional NeRFs, which use two successive MLPs.
We utilize the first MLP to estimate the density $\sigma$ and the expected distance $d$.
The second MLP predicts a two-channel feature map as in~\cite{guillard2022learning}:
Intensities $\textbf{i}$ and ray-drop probabilities $\textbf{p}$, respectively. Then, in the same way as for color, we can compute the per-view intensity and ray-drop probability by integrating along a ray $\textbf{r}$ as
$
  \hat{I}(\textbf{r}) = \sum_{i=1}^N T_i \big(1 - \exp(-\sigma_i \delta_i)\big)\textbf{i}_i\;,
  \hat{P}(\textbf{r}) = \sum_{i=1}^N T_i \big(1 - \exp(-\sigma_i \delta_i)\big)\textbf{p}_i \;.
$
Altogether, our \name{} can be formalized as a function $(\sigma, \textbf{i}, \textbf{p}) = f(\textbf{x}, \boldsymbol{\theta})$, and is summarized in \cref{fig:LiDAR-NeRF} (b).

\vspace{-5pt}
\mypara{Structural regularization.}
Despite the success in learning individual pixels, NeRFs tend to overlook local information, particularly in low-texture regions, leading to poor geometry, as evidenced in \cref{fig:reg}. Therefore, it is crucial to identify a suitable regularization technique to guide NeRF geometry. Notably, LiDAR point clouds exhibit clear patterns, and the transformed range images display significant structural features, which are essential for the network to learn, as illustrated in \cref{fig:LiDAR-NeRF} (c).
Initially, we attempted to apply the prevalent geometry regularization techniques, such as the smoothness-loss used in RegNeRF~\cite{niemeyer2022regnerf}
and the TV-loss used in Plenoxels~\cite{yu2021plenoxels}
, which aims to smooth neighboring points. However, we found that these techniques were not effective in large-scale scenes, where the differences between neighboring points can be significant. Subsequently, we explored learning structural information from the ground truth through the gradient loss. However, we observed that this approach was still insufficient, as the gradient loss was dominated by the rich-texture areas where the NeRF model excels. Consequently, we propose a novel structural regularization strategy based on the gradient loss, where we restrict regularization to low-texture areas, such as the ground.
Consequently, the structural regularization is defined as:
$
  \mathcal{L}_{\mathrm{reg}} = \begin{Vmatrix} \hat{G_M}(\mathbf{R}) - G_M(\mathbf{R}) \end{Vmatrix}_1 ,
  \label{eq:reg}
$
where $R$ is the set of training rays of local patches, and $G_M(\cdot)$ denotes the gradient operation
with low-texture areas mask.

\vspace{-6.5pt}
\mypara{Loss function.}
Our loss function includes four objectives
  {
    \begin{align}
      \mathcal{L}_{total} = \mathcal{L}_{\mathrm{distance}} +
      \lambda_1\mathcal{L}_{\mathrm{intensity}}(\mathbf{r})  + \lambda_2 \mathcal{L}_{\mathrm{raydrop}}(\mathbf{r}) +
      \lambda_3 \mathcal{L}_{\mathrm{reg}}\;,
    \end{align}
  }
with
$
  \mathcal{L}_{\mathrm{distance}}(\mathbf{r})=\sum_{\mathbf{r}\in R}\begin{Vmatrix} \hat{D}(\mathbf{r}) - D(\mathbf{r}) \end{Vmatrix}_1,
  \mathcal{L}_{\mathrm{intensity}}(\mathbf{r}) =\sum\begin{Vmatrix} \hat{I}(\mathbf{r}) - I(\mathbf{r}) \end{Vmatrix}_2^2,
  \mathcal{L}_{\mathrm{raydrop}}(\mathbf{r}) = \sum\begin{Vmatrix} \hat{P}(\mathbf{r}) - P(\mathbf{r}) \end{Vmatrix}_2^2,
$
where $R$ is the set of training rays,
and $\lambda$ are weight coefficients for each term.


\begin{figure}[t]
    \centering
    \vspace{-0.15cm}
    \includegraphics[width=0.92\linewidth]{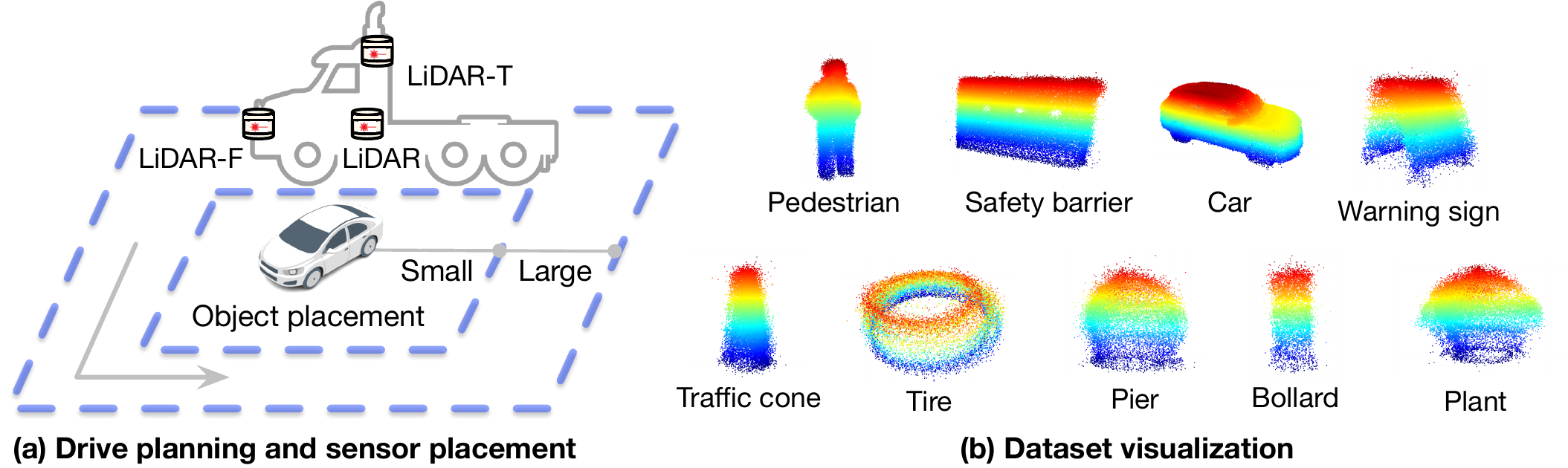}
    \caption{\textbf{(a)} We design two square paths of collection, small and large with 7 and 15 meters in length respectively.
        \textbf{(b)} Our \datasetname{} dataset encompasses 9 objects from common traffic categories. We align multiple frames here for better visualization.}
    \label{fig:dataset}
    \vspace{-10pt}
\end{figure}

\subsection{\datasetname{} Dataset}
\label{subsec:dataset}
As will be shown in our experiments, our approach can be applied to existing autonomous driving datasets~\cite{liao2022kitti360, caesar2020nuscenes, sun2020waymo} that have LiDAR sensors data. However, these datasets focus on scene-level LiDAR observations and thus depict views acquired from the vehicle driving along the scene, with fairly low diversity.
In other words, they lack the challenging diversity of object-centric data similar to that used for novel RGB view synthesis.
To facilitate future research in novel LiDAR view synthesis and verify the effectiveness of our \name{}, we therefore establish an object-centric multi-view LiDAR dataset, \datasetname{}, with carefully calibrated sensor poses and gathering multi-LiDAR sensor data from real autonomous vehicles.


\vspace{-6.5pt}
\mypara{Data collection.}
We collect the dataset in an enclosed area, employing self-driving vehicles with multiple LiDAR sensors.
The vehicles drive around the object in a square path twice, one large square and one small square, as shown in \cref{fig:dataset} (a).
To provide more diverse perspectives, we use various types of vehicles with different sensor placements and specifications. See \cref{tab:sensor}  of the appendix
for sensor details.

\vspace{-6.5pt}
\mypara{Data preparation.}
As shown in \cref{fig:dataset} (b), our \datasetname{} dataset consists of
nine objects from different common traffic categories.
After collecting multi-path, multi-sensor data, for each object, we crop out the region of interest, i.e., the object\footnote{The raw data will also be released to the community.}.
We carefully calibrate the LiDAR extrinsic parameters for every sensor, i.e., the relative location of the LiDAR to the ego body. The transformation matrix from the body coordinate system to the global world coordinate system is provided from the vehicle location based on GPS and IMU. Hence, in the dataset, we finally provide the calibration of the LiDAR to the global world, i.e., the \textit{lidar2world} matrix, to align all the frames.
Altogether, our \datasetname{} dataset contains more than 76k frames covering two types of collecting vehicles, three LiDAR settings, two collecting paths, and nine objects.
\vspace{-2pt}

\section{Experiments}
\vspace{-2pt}
We evaluate the scene-level and object-level synthesis ability of our \name{} both quantitatively and qualitatively. Additional results and details are provided in the supplementary material.

\vspace{-6.5pt}
\mypara{Baseline renderers.}
As generating novel LiDAR views remains unexplored, we moderately adapt existing model-based LiDAR simulators, i.e., LiDARsim~\cite{manivasagam2020lidarsim} and PCGen~\cite{li2022pcgen},
as the baseline renderers.
For exhaustive evaluation and comparisons, we also validate different settings of the baseline methods in \cref{subsec:baseline} and report the best value in the following sections.


\vspace{-6.5pt}
\mypara{Dataset.}
We conduct scene-scale experiments on the challenging KITTI-360~\cite{liao2022kitti360} dataset, which was collected in suburban areas.
We evaluate \name{} on LiDAR frames from 4 static suburb sequences as~\cite{fu2022panoptic}. Each sequence contains 64 frames, with 4 equidistant frames for evaluation.
We conduct the object-level experiments on our \datasetname{} dataset. For fast validation, we extract a pocket version of the dataset with only 7.3k frames covering the nine categories.

\vspace{-6.5pt}
\mypara{Metrics.}
For the novel LiDAR range images, we compute the usual metrics in depth estimation~\cite{godard2019digging}:
Root mean squared error (RMSE), and threshold accuracies ($\delta$1, $\delta$2, $\delta$3).
Moreover, we measure the structural quality  using the SSIM~\cite{wang2004ssim}.
To further evaluate the novel LiDAR view quality, we convert the rendered LiDAR range image to a point cloud
between the original and the novel point clouds $G_1, G_2$. It is computed as
  {\small
    {$
        \mathrm{C\mbox{-}D}\left(G_1, G_2\right)= \frac{1}{|G_1|} \sum_{x \in G_1} \min _{y \in G_2}\|x-y\|_2^2 +
        \frac{1}{|G_2|} \sum_{y \in G_2} \min _{x \in G_1}\|y-x\|_2^2 .
      $
    }
  }
We also report the F-Score between the two point clouds with a threshold of 5cm.
For the novel intensity image, it is evaluated using mean absolute error (MAE).

\begin{table*}[ht]
  \centering
  \caption{\textbf{Novel LiDAR view synthesis on scene-level KITTI-360 dataset and object-level \datasetname{} dataset}. \name{} outperforms the baseline in all metrics. Note that on the object-centric \datasetname{} with rich texture information, there is no need to apply structural regularization (SR).
  }
  \vspace{-0.2cm}
  \resizebox{1\textwidth}{!}{
    \addtolength{\tabcolsep}{-0.2pt}
    \begin{tabularx}{1\textwidth}{l|cccccccc}
      \toprule
      Method                                  & C-D$\downarrow$ & F-score$\uparrow$ & RMSE$\downarrow$ & $\delta$1$\uparrow$ & $\delta$2$\uparrow$ & $\delta$3$\uparrow$ & SSIM$\uparrow$ & MAE$\downarrow$ \\
      \midrule
      \multicolumn{3}{l}{\textit{\textbf{KITTI-360 dataset}}}                                                                                                                                               \\
      \midrule
      LiDARsim~\cite{manivasagam2020lidarsim} & 0.951           & 66.89             & 5.745            & 66.34               & 71.11               & 74.42               & 0.696          & 0.126           \\
      PCGen~\cite{li2022pcgen}                & 0.187           & 87.16             & 4.328            & 76.90               & 79.72               & 81.38               & 0.550          & 0.245           \\
      \midrule
      Ours-NeRF                               & 0.143           & 85.93             & 4.050            & 78.13               & 79.79               & 80.42               & 0.545          & 0.235           \\
      \textbf{Ours-iNGP (w/ SR)}              & \textbf{0.081}  & \textbf{92.49}    & \textbf{3.615}   & \textbf{82.18}      & \textbf{83.40}      & \textbf{83.97}      & \textbf{0.626} & \textbf{0.096}  \\

      \midrule
      \multicolumn{3}{l}{\textit{\textbf{\datasetname{} dataset}}}                                                                                                                                          \\
      \midrule

      LiDARsim~\cite{manivasagam2020lidarsim} & 0.022           & 96.01             & 5.984            & 83.43               & 83.43               & 83.43               & 0.612          & 4.143           \\
      PCGen~\cite{li2022pcgen}                & 0.078           & 90.40             & 7.558            & 73.13               & 73.13               & 73.13               & 0.217          & 6.268           \\
      \midrule
      Ours-NeRF                               & 0.028           & 92.81             & 3.864            & 93.65               & 93.65               & 93.65               & 0.462          & 2.642           \\
       \textbf{Ours-iNGP}        &  \textbf{0.005}	& \textbf{98.50} &	\textbf{1.305} &	\textbf{98.86} & \textbf{98.86}	& \textbf{98.86} &	\textbf{0.879} &	\textbf{1.057}  \\

      \bottomrule
    \end{tabularx}
  }
  \label{tab:results}
  \vspace{-7pt}
\end{table*}

%

\begin{figure*}[ht]
    \centering
    \vspace{-0.1cm}
    \includegraphics[width=1\linewidth]{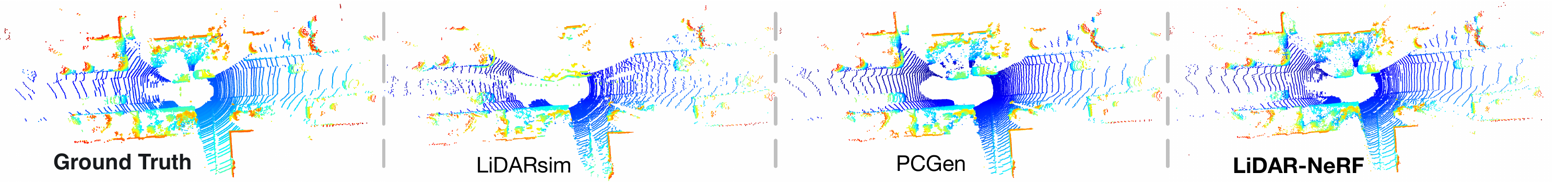}
    \caption{\textbf{Qualitative comparison on KITTI-360.}
        Our \name{} produces more realistic LiDAR patterns with highly detailed structure and geometry (zoom-in for the best of views).
    }
    \vspace{-8pt}
    \label{fig:kitti-res}
\end{figure*}

\subsection{Scene-level Synthesis}
\label{subsec:scene_exp}
We first evaluate the effectiveness of our \name{} on scene-level novel LiDAR view synthesis. The results are provided at the top of \cref{tab:results}.
Our \name{} significantly outperforms the baseline renderers over all metrics. To be specific, \name{} is superior to the baseline renderers with a comfortable margin in terms of C-D (0.081 vs 0.187, 0.951) and $\delta1$ (82.18 vs 76.90, 66.34).
In \cref{fig:kitti-res}, we provide qualitative results. Both methods are able to render general scene structures. While our \name{} produces more realistic LiDAR patterns and highly detailed structure and geometry.
The baseline LiDAR simulator mimics the physical LiDAR model through explicit 3D reconstruction and ray-tracing via traditional renderers.
As shown in \cref{fig:motivation},
the explicit 3D mesh reconstruction of LiDAR point clouds suffers from inaccuracy and tends to overestimate the object's surface. Consequently, these methods often produce unrealistic LiDAR patterns, as their explicit reconstruction and ray-casting neglect certain crucial features of LiDAR points. Specifically, the laser beams emitted by the LiDAR sensor can be affected by the surface material and normal direction, leading to the penetration of some beams through car glass and the loss of other beams. These effects are not fully considered in the explicit reconstruction process.


\subsection{Object-level Synthesis}
\label{subsec:obj_exp}

\begin{figure}[ht]
    \centering
    \vspace{-10pt}
    \includegraphics[width=0.85\linewidth]{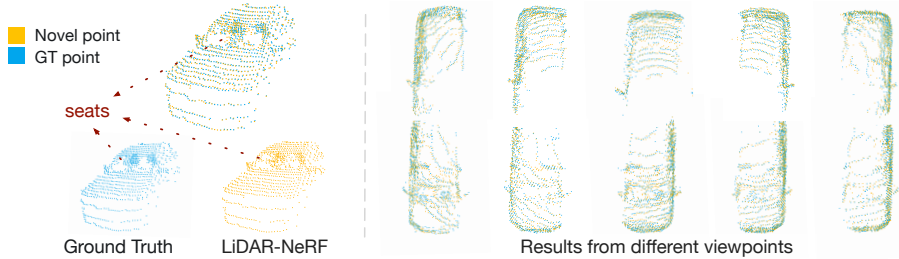}
    \vspace{0pt}
    \caption{\textbf{Qualitative results on \datasetname{} dataset}.
        \name{} can effectively encode 3D information and multiple attributes, enabling it to accurately model the behavior of beams as they penetrate car glass and reach seats.
        Moreover, the high quality of the results obtained from different viewpoints serves as compelling evidence of our method's effectiveness.
    }
    \vspace{-8pt}
    \label{fig:self-data-res}
\end{figure}

We conduct object-level synthesis experiments on the nine common traffic objects in our \datasetname{} dataset. As shown in \cref{tab:results},
our \name{} still significantly outperforms the baseline renderers by a large margin over all the metrics on all nine categories.
Furthermore, the qualitative visualization in \cref{fig:self-data-res} evidence that our approach produces significantly high-quality point clouds.


\subsection{Ablations}





\begin{table*}[ht]
  \vspace{-0.2cm}
  \centering
  \caption{\textbf{Ablations of our \name{}.} We ablate different architectures and regularization.}
  \vspace{-0.2cm}
  \addtolength{\tabcolsep}{0.8pt}
  \resizebox{1\textwidth}{!}{
    \begin{tabularx}{1\textwidth}{l|cccccccc}
      \toprule
      Component                           & C-D$\downarrow$ & F-score$\uparrow$ & RMSE$\downarrow$ & $\delta$1$\uparrow$ & $\delta$2$\uparrow$ & $\delta$3$\uparrow$ & SSIM$\uparrow$ & MAE$\downarrow$ \\
      \midrule

      \multicolumn{3}{l}{\textit{\textbf{Architecture}}}                                                                                                                                                \\
      \midrule
      w/ NeRF~\cite{mildenhall2021nerf}   & 0.126           & 87.64             & 3.948            & 78.26               & 79.57               & 80.09               & 0.555          & 0.226           \\
      w/ iNGP~\cite{muller2022instantNGP} & 0.088           & 92.00             & 3.577            & 80.40               & 81.47               & 81.87               & 0.605          & 0.101           \\

      \midrule
      \multicolumn{3}{l}{\textit{\textbf{Regularization (w/ iNGP)}}}                                                                                                                                    \\
      \midrule
      w/o Reg                             & 0.088           & 92.00             & 3.577            & 80.40               & 81.47               & 81.87               & 0.605          & 0.101           \\
      Smooth loss                         & 0.085           & 92.91             & 3.576            & 80.48               & 81.50               & 81.92               & 0.606          & 0.104           \\
      Gradient loss                       & 0.080           & 92.91             & 3.601            & 80.67               & 81.65               & 82.06               & 0.607          & 0.103           \\

      \textbf{Struc-Reg}                  & \textbf{0.077}  & \textbf{92.98}    & \textbf{3.511}   & \textbf{82.25}      & \textbf{83.28}      & \textbf{83.73}      & \textbf{0.635} & \textbf{0.096}  \\

      \bottomrule
    \end{tabularx}
  }
  \label{tab:ablation_ref}
  \vspace{-9pt}
\end{table*}

\begin{figure*}[ht]
    \centering
    \vspace{-0.1cm}
    \includegraphics[width=1\linewidth]{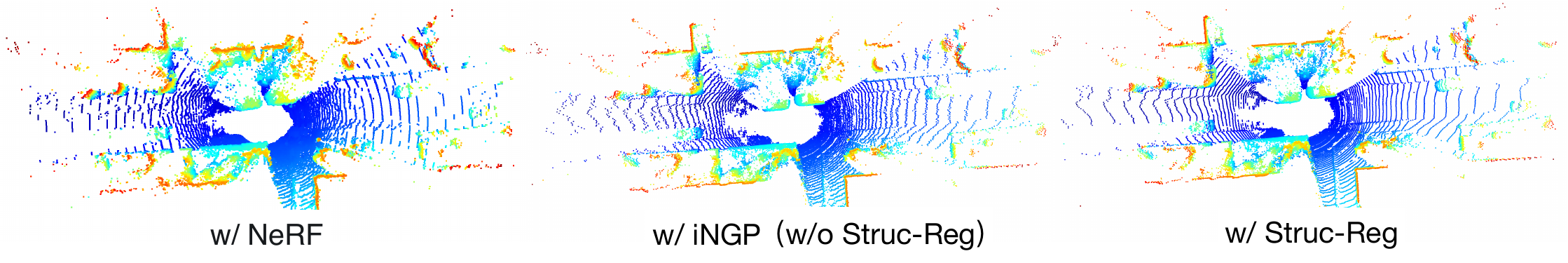}
    \vspace{-0.4cm}
    \caption{
        \textbf{Qualitative evaluation of various configurations.}
        The iNGP's hybrid 3D grid architecture achieves more detailed structures. Our structural regularization significantly improves the geometry estimation and produces more realistic LiDAR patterns.
    }
    \vspace{-8pt}
    \label{fig:reg}
\end{figure*}

Our investigation into various configurations within the \name{} framework includes an exploration of different architectures and regularization. As shown in \cref{tab:ablation_ref} and \cref{fig:reg}, we examine the widely used NeRF~\cite{mildenhall2021nerf} and the best performing Instance-NGP (iNGP)~\cite{muller2022instantNGP}. iNGP introduces a hybrid 3D grid structure with a multi-resolution hash encoding and lightweight MLPs that is more expressive than the vanilla NeRF and achieves better performance. Thus we chose to use iNGP as our base architecture.
Additionally, we compare the structural regularization (Struc-Reg) of our \name{} with the aforementioned regularization in \cref{subsec:lidar-nerf}. The results in \cref{tab:ablation_ref} and \cref{fig:reg} demonstrate the effectiveness of our structural regularization both quantitatively and qualitatively.

\section{Conclusion}
We have introduced the new task of novel LiDAR view synthesis and proposed the first differentiable LiDAR renderer. Our proposed method, \name{}, jointly learns the geometry and attributes of 3D points with structural regularization, resulting in more accurate and realistic LiDAR patterns. We further established the \datasetname{} dataset, which contains 9 objects over 360-degree LiDAR viewpoints acquired with multiple sensors. Our experiments on both scene-level and our object-level data have evidenced the superiority of our approach over model-based simulators. Importantly, our approach is simple and does not rely on explicit 3D reconstruction and rendering engines. We hope that our work can shed light on novel LiDAR view synthesis and inspire future research in this field.

\clearpage

{\small
    \bibliography{egbib}

}

\clearpage
\appendix
\section{Limitations and Future work}
As this paper is the first attempt at novel LiDAR view synthesis, there remains room for improvement.
Our \name{} draws inspiration from the original NeRF formalism. As such, it is better suited to static scenes, and requires per-scene optimization.
Fortunately, much progress is being made in handling dynamic scenes~\cite{pumarola2021d-nerf, ost2021NSG, jang2022d-TensoRF} and generalization~\cite{yu2021pixelnerf, trevithick2021grf} in image-based NeRF, and we expect these advances to facilitate the development of counterparts for novel LiDAR view synthesis.
Note that, in this work, we have considered the problem of synthesizing LiDAR data only, but jointly rendering LiDAR and images is a natural step forward. We therefore plan to extend our \datasetname{} dataset to a multimodal one.
Moreover, we are integrating various LiDAR renderers, including LiDAR simulators and our proposed LiDAR-NeRF framework, with the goal of developing a unified Codebase for novel LiDAR view synthesis that can benefit the community.
Altogether, we hope that our work will inspire other researchers to contribute to the development of novel LiDAR view synthesis.

\section{Additional Details}
\label{sec:details}
\subsection{Ablation details}
All ablation experiments were conducted on the seq-1908-1971 of the KITTI-360 dataset, which is a large, clear scene with numerous objects, making it an ideal sequence for comparison.

\subsection{\name{} details}
\vspace{-6pt}
\mypara{NeRF revisited.}
NeRF represents a scene as a continuous volumetric radiance field. For a given 3D point $\textbf{x} \in \mathbb{R}^3$ and a
viewing direction $\boldsymbol{\theta}$, NeRF learns an implicit function $f$ that estimates the differential density $\sigma$ and view-dependent RGB color $\textbf{c}$ as $(\sigma, \textbf{c}) = f(\textbf{x}, \boldsymbol{\theta})$.
Specifically, NeRF uses volumetric rendering to render image pixels.
Given a pose $\mathbf{P}$, it casts rays $\mathbf{r}$ originating from $\mathbf{P}$'s center of projection $\mathbf{o}$ in direction $\mathbf{d}$, i.e., $\mathbf{r}(t) = \mathbf{o} + t\mathbf{d}$.
The implicit radiance field is then integrated along this ray, and the color is approximated by integrating over samples lying along the ray. This is expressed as
$
  \hat{C}(\textbf{r}) = \sum_{i=1}^N T_i \big(1 - \exp(-\sigma_i \delta_i)\big)\textbf{c}_i \;,
  \label{eq:nerf}
$
where $ T_i = \exp \left(-\sum_{j=1}^{i-1} \sigma_j \delta_j \right)$ indicates the accumulated transmittance along ray $\textbf{r}$ to the sampled point $t_i$,
$\textbf{c}_i$ and $\sigma_i$ are the corresponding color and density at $t_i$, and $\delta_i = t_{i+1} - t_i$ is the distance between adjacent samples.
Moreover, per-view depth $\hat{D}(\mathbf{r})$ can also be approximated by calculating the expectation of the samples along the ray, i.e.,
$
  \hat{D}(\textbf{r}) = \sum_{i=1}^N T_i \big(1 - \exp(-\sigma_i \delta_i)\big)t_i \;.
  \label{eq:depth}
$

\mypara{\name{} (w iNGP).}
Our \name{} is implemented based on torch-ngp~\cite{torch-ngp}, which introduces a hybrid 3D grid structure with a multi-resolution hash encoding and lightweight MLPs.
We optimize our \name{} model per scene using a single NVIDIA GeForce RTX 3090 GPU. For each scene,
we center the LiDAR point clouds by subtracting the origin of the global world coordinate system from the scene's central frame.
Then, the scene frames are scaled by a factor such that the region of interest falls within a unit cube, which is required by most positional encodings used in NeRFs.
We use Adam~\cite{kingma2014adam} with a learning rate of 1e-2 to train our models. The coarse and fine networks are sampled 768 and 64 samples per ray, respectively. The finest resolution of the hash encoding is set to 32768.
For structural regularization, we employ patch-wise training with a patch size of 2x8 and mask gradients smaller than the threshold of 0.1. The optimization process consists of a total of 30k steps, with $\lambda_1 = 1$, $\lambda_2 = 1$, and $\lambda_3 = 1e2$.
For our \datasetname{} dataset, we first get the 3D box of each object, and then project to the range view.
Only a few rays within the box are trained, so the network converges quickly.

\mypara{\name{} (w NeRF).}
\label{subsec:nerf_implementation}
The \name{} (w NeRF) is implemented based on nerf-pytorch~\cite{lin2020nerfpytorch}. The coarse and fine networks are sampled 64 and 128 times, respectively, during training. The highest frequency of the coordinates is set to $2^{15}$.
We use Adam~\cite{kingma2014adam} with a learning rate of 5e-4 to train our models. We set the loss weights to $\lambda_1$ = 1, $\lambda_2$ = 1, and optimize the total loss $\mathcal{L}_{total}$ for 400k iterations with a batch size of 2048.

\subsection{\datasetname{} details}
\begin{table}[h!]
\centering
\vspace{-17pt}
\caption{
\textbf{LiDAR sensor configurations.}
}
\begin{tabularx}{0.98\textwidth}{p{1.5cm} | p{11.5cm}}
\toprule
\textbf{Sensor} &   \textbf{Details}\\
\midrule
LiDAR LiDAR-F &   Spinning, $64$ beams,  $10\text{Hz}$ capture frequency, \ang{360} horizontal FOV, \ang{0.6} horizontal resolution, \ang{-52.1} to \ang{+52.1} vertical FOV, $\leq 60m$ range, $\pm 3\text{cm}$ accuracy.\\
\midrule
LiDAR-T   &   Spinning, $64$ beams, $20\text{Hz}$ capture frequency, \ang{360} horizontal FOV, \ang{0.4} horizontal resolution, \ang{-25} to \ang{+15} vertical FOV, $\leq 200m$ range, $\pm 2\text{cm}$ accuracy.\\
\bottomrule
\multicolumn{2}{l}{{Sensor location: F: front. T: top.} }
\end{tabularx}
\vspace{-10pt}

\label{tab:sensor}
\end{table}
We provide detailed sensor specifications in \cref{tab:sensor}.

\subsection{Baseline details}
\label{subsec:baseline}
\begin{figure*}[h!]
    \centering
    \includegraphics[width=0.95\textwidth]{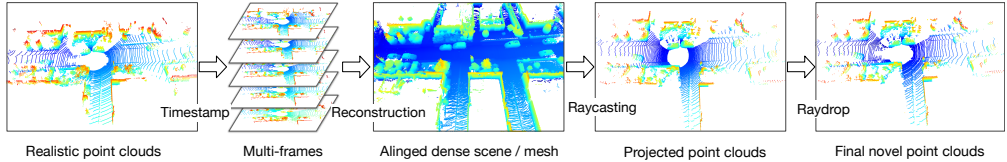}
    \caption{\textbf{Baseline multi-step LiDAR simulator pipeline.} }
    \label{fig:LiDAR-Render}
\end{figure*}

As generating novel LiDAR views remains unexplored, we moderately adapt existing model-based LiDAR simulators as our baseline.
Given a sequence of LiDAR frames, the objective is to generate novel LiDAR scenes with realistic geometry. In essence, the baseline, summarized in \cref{fig:LiDAR-Render}, follows the physics-based, multi-step approach of existing simulation pipelines~\cite{manivasagam2020lidarsim, li2022pcgen}.
%
In short, they first gather a set of LiDAR frames, which are transformed into the global world coordinate system. The resulting aligned dense 3D scene or mesh is projected to the novel view via ray-casting. Finally, the problem of LiDAR ray-dropping is simulated to improve realism.
Because the official codes of LiDARsim~\cite{manivasagam2020lidarsim} and PCGen~\cite{li2022pcgen} are not publicly available, we re-implemented them following the papers as closely as possible.
Below, we discuss the implementation in more detail.

\subsubsection{LiDARsim}

\mypara{Point cloud meshing and mesh-ray intersection.} Following LiDARsim~\cite{manivasagam2020lidarsim}, our pipeline also consists of two steps: point cloud meshing and mesh-ray intersection. The point cloud meshing step is implemented based on the Poisson surface reconstruction algorithm~\cite{kazhdan2006poisson}, which is a modification from the original author's Surfel-based meshing. The mesh-ray intersection step is implemented based on Intel Embree using the Open3D library~\cite{zhou2018open3d}. The performance of Poisson surface reconstruction is sensitive to hyperparameters, namely the maximum depth of the tree and the weight threshold for selecting mesh vertices. For each dataset, we perform grid search of these two parameters on the training set and select the best-performing one on the test set.

\mypara{Ray-drop network.} To train the ray-drop network, following LiDARsim~\cite{manivasagam2020lidarsim}, we concatenate range, intensity, incidence angle (in cosine), and the normal of surface hit as inputs. We then train a U-Net with 4 down-sampling (with 64, 128, 256, and 512 channels respectively) and 4 up-sampling layers (with 1024, 512, 256, and 128 channels respectively) to predict the ray-drop mask. We train on the training frames for 10 epochs. The final range image prediction is the ray-casted results multiplied by the ray-drop mask. We use this final range image for evaluation.

\subsubsection{PCGen}
\mypara{Ray-casting.} Ray-casting defines the intersections between the laser beams and the dense point cloud. The standard approach is closest-point (CP) ray-casting~\cite{langer2020domain}, which projects the dense point cloud to a range image using \cref{eq:rangeview}, and, for each ray, selects the point with the smallest measured distance, as is commonly done in rendering pipelines with the so-called z-buffer.
In real-world scenarios, however, the calibration, sensor synchronization, and other properties~\cite{yu2022benchmarking} are affected by different noise sources, and the points in the dense point cloud do not strictly lie on the scene surface. Thus, considering only the closest point tends to render noisy ray-casted point clouds and more points in the z-buffer need to be taken into account. PCGen~\cite{li2022pcgen} propose the first peak averaging (FPA) raycasting, averaging the points within a certain threshold near the closest point, and weighting them by their inverse distance.

\mypara{Ray-drop network.} Ray-casting yields a nearly perfect point cloud in the novel view. However, the laser returns of a real LiDAR sensor are affected by many factors, such as the distance to the scene, the incidence angle, and the material texture and reflectivity.
To improve realism, we follow from PCGen~\cite{li2022pcgen} to employ a small surrogate MLP to learn the ray-drop, which we dub ray-drop MLP.
Specifically, we model a ray's return probability as
$
  p = MLP(\boldsymbol\theta',d,i),
  \label{eq:raydrop}
$
where $\boldsymbol\theta'$, $d$, and $i$ are the viewing direction in the local coordinate system, the distance, and the intensity of the ray.
For the implementation of ray-drop MLP, it has 4 layers with a width of 128. It is trained with Adam~\cite{kingma2014adam} with a learning rate of 5e-3, and supervised with MSE loss for 10k iterations with a batch size of 2048.


\section{Additional Experimental Results}
\subsection{More results.}
\begin{table*}[h!]
  \centering
  \caption{\textbf{More results of novel LiDAR view synthesis on KITTI-360}. \name{} outperforms the baseline over all sequences in all metrics.}
  \addtolength{\tabcolsep}{0.8pt}
  \resizebox{1\textwidth}{!}{
    \begin{tabularx}{1\textwidth}{l|cccccccc}
      \toprule
      Sequence       & C-D$\downarrow$ & F-score$\uparrow$ & RMSE$\downarrow$ & $\delta$1$\uparrow$ & $\delta$2$\uparrow$ & $\delta$3$\uparrow$ & SSIM$\uparrow$ & MAE$\downarrow$ \\
      \midrule
      \multicolumn{3}{l}{\textit{\textbf{LiDARsim~\cite{manivasagam2020lidarsim}}}}                                                                                                \\
      \midrule
      Seq 1538--1601 & 0.794           & 72.13             & 5.455            & 67.49               & 73.68               & 78.17               & 0.719          & 0.125           \\
      Seq 1728--1791 & 0.819           & 65.38             & 6.083            & 67.20               & 70.86               & 74.11               & 0.694          & 0.122           \\
      Seq 1908--1971 & 0.981           & 66.77             & 5.404            & 67.33               & 72.70               & 74.76               & 0.711          & 0.122           \\
      Seq 3353--3416 & 1.211           & 63.29             & 6.038            & 63.35               & 67.20               & 70.64               & 0.660          & 0.133           \\
      \midrule
      Average        & 0.951           & 66.89             & 5.745            & 66.34               & 71.11               & 74.42               & 0.696          & 0.126           \\

      \midrule \midrule
      \multicolumn{3}{l}{\textit{\textbf{PCGen~\cite{li2022pcgen}}}}                                                                                                               \\
      \midrule
      Seq 1538--1601 & 0.159           & 88.64             & 4.091            & 75.64               & 79.62               & 82.08               & 0.558          & 0.254           \\
      Seq 1728--1791 & 0.194           & 83.87             & 4.560            & 76.19               & 79.37               & 81.35               & 0.545          & 0.243           \\
      Seq 1908--1971 & 0.220           & 87.75             & 4.029            & 77.35               & 79.95               & 81.05               & 0.559          & 0.238           \\
      Seq 3353--3416 & 0.173           & 88.37             & 4.633            & 78.40               & 79.92               & 81.05               & 0.536          & 0.244           \\

      \midrule
      Average        & 0.187           & 87.16             & 4.328            & 76.90               & 79.72               & 81.38               & 0.550          & 0.245           \\

      \midrule
      \midrule
      \multicolumn{3}{l}{\textit{\textbf{\name{} (w/ NeRF)}}}                                                                                                                      \\
      \midrule
      Seq 1538--1601 & 0.148           & 84.88             & 4.007            & 76.66               & 79.09               & 79.85               & 0.527          & 0.242           \\
      Seq 1728--1791 & 0.148           & 83.88             & 4.207            & 78.44               & 80.35               & 81.09               & 0.549          & 0.239           \\
      Seq 1908--1971 & 0.126           & 87.64             & 3.948            & 78.26               & 79.57               & 80.09               & 0.555          & 0.226           \\
      Seq 3353--3416 & 0.150           & 87.31             & 4.039            & 79.15               & 80.14               & 80.63               & 0.550          & 0.233           \\
      \midrule
      Average        & 0.143           & 85.93             & 4.050            & 78.13               & 79.79               & 80.42               & 0.545          & 0.235           \\

      \midrule\midrule
      \multicolumn{5}{l}{\textit{\textbf{\name{} (w/ iNGP and StrucReg)}}}                                                                                                         \\
      \midrule
      Seq 1538--1601 & 0.073           & 92.55             & 3.530            & 80.76               & 82.71               & 83.52               & 0.597          & 0.102           \\
      Seq 1728--1791 & 0.088           & 90.95             & 3.766            & 82.91               & 84.26               & 84.91               & 0.646          & 0.091           \\
      Seq 1908--1971 & 0.077           & 92.98             & 3.511            & 82.25               & 83.28               & 83.73               & 0.635          & 0.096           \\
      Seq 3353--3416 & 0.086           & 93.46             & 3.654            & 82.78               & 83.36               & 83.71               & 0.625          & 0.094           \\
      \midrule
      Average        & 0.081           & 92.49             & 3.615            & 82.18               & 83.40               & 83.97               & 0.626          & 0.096           \\

      \bottomrule
    \end{tabularx}
  }
  \label{tab:seqs}
\end{table*}

\mypara{KITTI-360 dataset.}
We report detailed results on the four sequences of the KITTI-360 dataset in \cref{tab:seqs}.
Our \name{} consistently outperforms the baseline over all sequences in all metrics.



\begin{table*}[h!]
  \centering
  \caption{\textbf{Novel LiDAR view synthesis on \datasetname{}}. Our \name{} outperforms the baseline in all metrics.
  }
  \vspace{-0.2cm}
  \addtolength{\tabcolsep}{0.8pt}
  \resizebox{1\textwidth}{!}{
    \begin{tabularx}{1\textwidth}{l|cccccccc}
      \toprule
      Object-Category & C-D$\downarrow$ & F-score$\uparrow$ & RMSE$\downarrow$ & $\delta$1$\uparrow$ & $\delta$2$\uparrow$ & $\delta$3$\uparrow$ & SSIM$\uparrow$ & MAE$\downarrow$ \\
      \midrule \midrule
      \multicolumn{3}{l}{\textit{\textbf{LiDARsim~\cite{manivasagam2020lidarsim} }}}                                                                                                \\
      \midrule
      Bollard         & 0.011           & 98.92             & 5.751            & 84.64               & 84.64               & 84.64               & 0.518          & 2.172           \\
      Car             & 0.067           & 94.06             & 6.331            & 82.35               & 82.37               & 82.37               & 0.645          & 1.426           \\
      Pedestrian      & 0.011           & 97.91             & 4.788            & 90.20               & 90.20               & 90.20               & 0.748          & 4.353           \\
      Pier            & 0.016           & 95.33             & 6.606            & 78.67               & 78.67               & 78.67               & 0.510          & 2.541           \\
      Plant           & 0.018           & 93.81             & 5.315            & 86.45               & 86.45               & 86.45               & 0.667          & 0.852           \\
      Safety barrier  & 0.015           & 97.95             & 6.697            & 78.29               & 78.29               & 78.29               & 0.565          & 5.801           \\
      Tire            & 0.021           & 95.45             & 6.936            & 75.53               & 75.53               & 75.53               & 0.566          & 0.457           \\
      Traffic cone    & 0.013           & 99.59             & 5.861            & 88.14               & 88.14               & 88.14               & 0.628          & 19.14           \\
      Warning sign    & 0.024           & 91.05             & 5.569            & 86.57               & 86.57               & 86.57               & 0.663          & 0.546           \\
      \midrule
      Average         & 0.022           & 96.01             & 5.984            & 83.43               & 83.43               & 83.43               & 0.612          & 4.143           \\

      \midrule \midrule
      \multicolumn{3}{l}{\textit{\textbf{PCGen~\cite{li2022pcgen}}}}                                                                                                                \\
      \midrule

      Bollard         & 0.021           & 96.71             & 9.928            & 57.43               & 57.43               & 57.43               & 0.082          & 4.046           \\
      Car             & 0.446           & 72.23             & 5.829            & 86.37               & 86.37               & 86.37               & 0.331          & 0.123           \\
      Pedestrian      & 0.063           & 92.43             & 4.364            & 92.61               & 92.61               & 92.61               & 0.472          & 6.073           \\
      Pier            & 0.017           & 96.92             & 9.534            & 58.11               & 58.11               & 58.11               & 0.088          & 4.557           \\
      Plant           & 0.021           & 91.33             & 6.729            & 76.26               & 76.26               & 76.26               & 0.280          & 1.475           \\
      Safety barrier  & 0.070           & 74.88             & 5.382            & 87.33               & 87.33               & 87.33               & 0.207          & 5.689           \\
      Tire            & 0.026           & 94.84             & 8.748            & 62.58               & 62.58               & 62.58               & 0.095          & 0.714           \\
      Traffic cone    & 0.014           & 98.52             & 8.698            & 71.80               & 71.80               & 71.80               & 0.223          & 32.26           \\
      Warning sign    & 0.020           & 95.77             & 8.810            & 65.72               & 65.72               & 65.72               & 0.171          & 1.472           \\
      \midrule
      Average         & 0.078           & 90.40             & 7.558            & 73.13               & 73.13               & 73.13               & 0.217          & 6.268           \\

      \midrule \midrule
      \multicolumn{3}{l}{\textit{\textbf{\name{} (w/ NeRF)}}}                                                                                                                       \\
      \midrule
      Bollard         & 0.028           & 90.48             & 3.805            & 93.56               & 93.56               & 93.56               & 0.245          & 0.426           \\
      Car             & 0.033           & 92.77             & 4.044            & 93.14               & 93.14               & 93.14               & 0.544          & 0.457           \\
      Pedestrian      & 0.018           & 96.33             & 3.299            & 95.70               & 95.70               & 95.70               & 0.627          & 6.454           \\
      Pier            & 0.014           & 96.68             & 3.701            & 93.52               & 93.52               & 93.52               & 0.353          & 0.837           \\
      Plant           & 0.019           & 96.30             & 3.349            & 95.02               & 95.02               & 95.02               & 0.571          & 0.472           \\
      Safety barrier  & 0.045           & 89.59             & 3.693            & 93.87               & 93.87               & 93.87               & 0.517          & 4.070           \\
      Tire            & 0.039           & 89.52             & 3.640            & 93.27               & 93.27               & 93.27               & 0.448          & 0.100           \\
      Traffic cone    & 0.026           & 92.72             & 4.747            & 92.66               & 92.66               & 92.66               & 0.428          & 10.88           \\
      Warning sign    & 0.033           & 90.87             & 4.296            & 92.62               & 92.62               & 92.62               & 0.424          & 0.082           \\
      \midrule
      Average         & 0.028           & 92.81             & 3.864            & 93.65               & 93.65               & 93.65               & 0.462          & 2.642           \\

      \midrule \midrule
      \multicolumn{3}{l}{\textit{\textbf{\name{} (w/ iNGP)}}}                                                                                                                       \\
      \midrule

      Bollard  & 0.007 &  98.54 &  0.974 &  99.13 &  99.13 &  99.13  & 0.786 &  0.723 \\
      Car  &  0.005 & 99.33 &    2.256 & 97.85 & 97.85 & 97.85 & 0.842 &   0.436    \\
      Pedestrian & 0.001 & 99.97 & 1.381& 99.24 & 99.24& 99.24 &0.941& 1.650    \\
      Pier    &   0.004 & 98.14  & 1.047 & 99.08 & 99.08 & 99.08 & 0.889 &0.671 \\
      Plant  &  0.001 & 99.36 & 0.415 & 99.80 & 99.80& 99.80 & 0.976 &0.464   \\
      Safety barrier & 0.018 & 92.41& 2.624& 96.63 & 96.63& 96.63 & 0.622 &2.640        \\
      Tire  &   0.001 &100.0 &  0.563& 99.64& 99.64& 99.64 &0.965& 0.174     \\
      Traffic cone & 0.002 & 100.0 & 1.221&  99.32&  99.32&  99.32& 0.949 & 2.493    \\
      Warning sign & 0.006 & 98.77& 1.266 &99.13&99.13&99.13 &0.948& 0.259 \\

      \midrule
      Average  &  0.005	& 98.50 &	1.305 &	98.86 & 98.86	& 98.86 &	0.879 &	1.057     \\

      \bottomrule
    \end{tabularx}
  }
  \label{tab:exp_selfdata_detail}
  \vspace{-4pt}
\end{table*}

\mypara{\datasetname{} dataset.}
We report detailed results on nine object categories of \datasetname{} dataset in \cref{tab:exp_selfdata_detail}. Our
\name{} consistently outperforms the baseline over all categories in all metrics.

\subsection{More ablations of \name{}.}

\begin{table*}[h!]
     \centering
     \caption{\textbf{Ablation of \name{} (w NeRF) training strategies.} }
         \addtolength{\tabcolsep}{-1.1pt}
         \resizebox{1\textwidth}{!}{
      \begin{tabularx}{1\textwidth}{ccc|cccccccc}
        \toprule
 Scale & Contract & Cos-lr & C-D$\downarrow$  & F-score$\uparrow$    & RMSE$\downarrow$ & $\delta$1$\uparrow$ & $\delta$2$\uparrow$       & $\delta$3$\uparrow$   & SSIM$\uparrow$ & MAE$\downarrow$\\
\midrule 

\colorbox{mycyan}{\xmark} & \xmarkg & \cmarkg  & 132.4 & 00.19 & 9.910 & 00.07 & 00.62 & 2.087 & 0.296 & 0.491 \\
\xmarkg & \colorbox{mycyan}{\cmark} & \cmarkg & 0.129 & 87.08 &  3.849 &  79.63 & 81.09 & 81.64 &   0.575 & 0.231 \\

 \cmarkg & \xmarkg  & \colorbox{mycyan}{\xmark}  & 0.146 & 84.07 & 3.978 & 77.81 & 79.33 & 79.81 &  0.549 & 0.225\\

 \cmark & \xmark & \cmark & 0.126 &  87.64 & 3.948 &  78.26 & 79.57 & 80.09& 0.555 & 0.226 \\

\bottomrule
      \end{tabularx}
      }
     \label{tab:ablation_nerf}
   \end{table*}

\mypara{\name{} (w NeRF).} We ablate the training strategies of our \name{} (w NeRF) in \cref{tab:ablation_nerf}. In the training process, we scale (Scale) the scene frames with a factor such that the region of interest falls within a unit cube. There is also a parameterization function (contract) used in~\cite{barron2021mip, xie2023s-nerf} as:
\begin{align}
  \text{contract}(x) & = \left\{
  \begin{array}{cc}
    x/r,                                   &
    \text{if}~~\|x\| \leq r,                                   \\
    (1+b-\frac{br}{\|x\|})\frac{x}{\|x\|}, & \text{otherwise.}
  \end{array}
  \right.
\end{align}
Where $r$ and $b$ are the radius parameters to decide the mapping boundary. We set $r=10$ and $b=1$ for the scene point clouds.
As expected, without the parameterization function, the model hardly learned anything. While the 'Scale' and 'Contract' functions achieve comparable results.
Moreover, we employ a cosine schedule with one thousand iterations to warm up the learning rate (Cos-lr) to stable training, which slightly improves the performance.

\begin{table*}[h!]
  \addtolength{\tabcolsep}{-3.4pt}
  \centering
  \caption{\textbf{Ablation of baseline simulators. }}
  \resizebox{1\textwidth}{!}{
    \begin{tabularx}{1\textwidth}{ll|cccccccc}
      \toprule
      Method                                    & Component                  & C-D$\downarrow$ & F-score$\uparrow$ & RMSE$\downarrow$ & $\delta$1$\uparrow$ & $\delta$2$\uparrow$ & $\delta$3$\uparrow$ & SSIM$\uparrow$ & MAE$\downarrow$ \\
      \midrule

      \multirow{2}{*}{LiDARsim~\cite{manivasagam2020lidarsim}}
                                                & Mesh ray-casting           & 1.026           & 64.62             & 5.674            & 52.09               & 57.90               & 60.15               & 0.592          & 0.156           \\
                                                & \texttt{+} Ray-drop U-Net  & 0.981           & 66.77             & 5.404            & 67.33               & 72.70               & 74.76               & 0.711          & 0.122           \\
      \midrule

      \multirow{3}{*}{PCGen~\cite{li2022pcgen}} & CP ray-casting             & 0.259           & 83.87             & 4.312            & 65.06               & 67.86               & 69.09               & 0.153          & 0.274           \\

                                                & \texttt{+} FPA ray-casting & 0.248           & 85.43             & 4.332            & 65.21               & 67.93               & 69.13               & 0.155          & 0.280           \\

                                                & \texttt{+} Ray-drop MLP    & 0.220           & 87.75             & 4.029            & 77.35               & 79.95               & 81.05               & 0.220          & 0.238           \\

      \bottomrule
    \end{tabularx}
  }
  \label{tab:ablation_baseline}
\end{table*}

\subsection{More ablations of baseline.}
For exhaustive evaluation and fair comparisons, we also validate different settings of the baseline methods as follows.
In \cref{tab:ablation_baseline}, we investigate the different components of baseline simulators.
\mypara{LiDARsim.} For LiDARsim~\cite{manivasagam2020lidarsim}, the ray-drop U-Net can both boost the performance and demonstrate its effectiveness.

\mypara{PCGen.} For PCGen~\cite{li2022pcgen}, the FPA z-buffer ray-casting and ray-drop MLP can both boost the performance and demonstrate their effectiveness.

It is worth noting that training the ray-drop network necessitates considerable effort, entailing the initial rendering of training sets, followed by the construction of paired training sets for the ray-drop network, comprising the rendered outcomes and their corresponding ground truth. Subsequently, it is imperative to meticulously adjust the model's architecture and training parameters to achieve superior results for each scene or object.


\section{Additional Qualitative Visualization.}
\begin{figure*}[h!]
    \centering
    \includegraphics[width=1\linewidth]{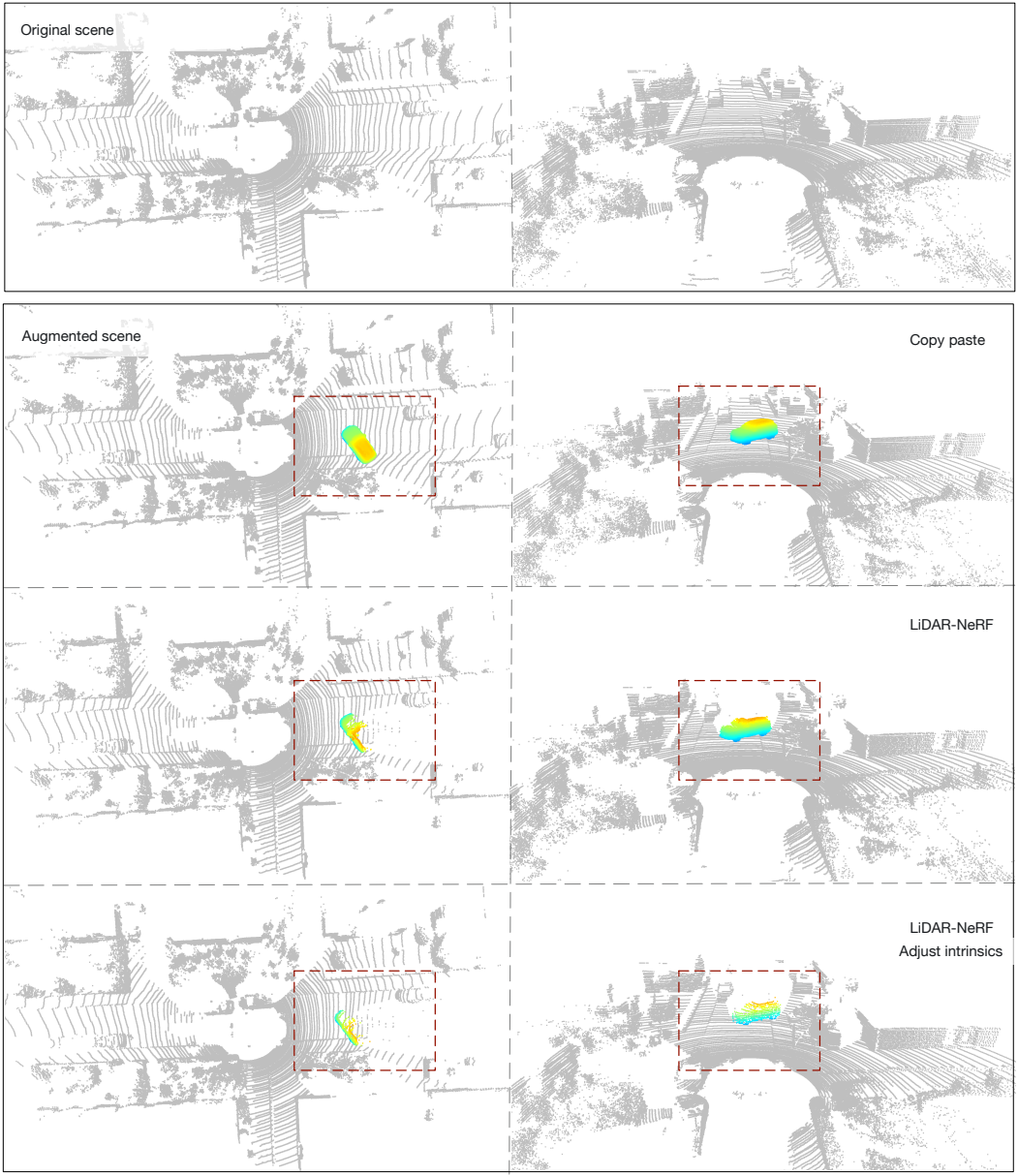}
    \caption{\textbf{Scene editing.} The augmented scene from our \name{} has realistic occlusion effects and consistent LiDAR pattern thanks to our pseudo range-image formalism, compared with the common cope-paste strategy. }

    \label{fig:scene-edit_paste}
\end{figure*}

\mypara{Scene editing.}
\label{subsec:scene_edit}
As our \name{} can effectively synthesize novel LiDAR views at both scene level and object level, it can be exploited to achieve scene editing. We provide an example for novel scene arrangements, which corresponds to editing the scene from the KITTI-360 dataset by fusing novel objects from our \datasetname{} dataset. Given the 6D pose (3D translation and yaw, pitch, and roll rotations) of the new object, we first render the corresponding novel view of the object, and then paste it to the desired position in the scene.
Furthermore, it is worth mentioning that our method has the capability to
adjust the intrinsics of LiDAR, thereby addressing the issue of inconsistent LiDAR patterns resulting from the use of different LiDAR devices in the \datasetname{} and KITTI-360 datasets.
As illustrated in \cref{fig:scene-edit_paste}, our \name{} can render the corresponding novel view, and the yield augmented scene has realistic occlusion effects and a consistent LiDAR pattern, compared with the common cope-paste strategy. We provide more visualizations in the video in the supplementary material.

\begin{figure*}[ht]
    \centering
    \includegraphics[width=1\linewidth]{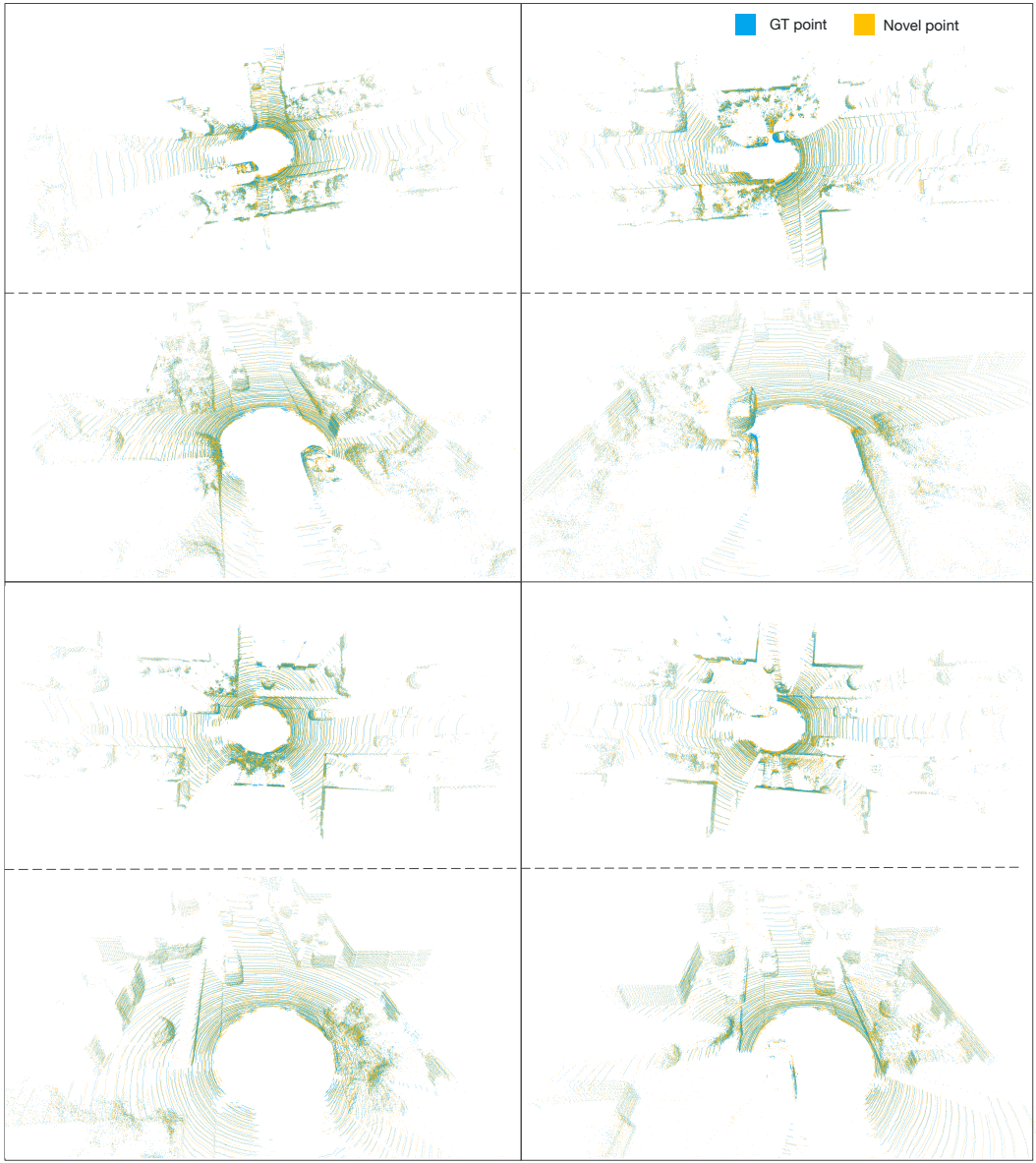}
    \caption{\textbf{Qualitative results on KITTI-360.} The high quality of the results from different view-points demonstrates the effectiveness of our method.}
    \label{fig:kitti-data-res}
\end{figure*}

\mypara{Qualitative results on KITTI-360.}
We provide more qualitative results on KITTI-360 dataset in \cref{fig:kitti-data-res}, which shows that
our \name{} produces high-quality point clouds fidelity with the ground truth.

\mypara{Video demo.} In addition to the figures, we have attached a video demo in the supplementary materials, which consists of hundreds of frames that provide a more comprehensive evaluation of our proposed approach.

\end{document}